\documentclass[sn-mathphys,Numbered]{sn-jnl}% Math and Physical Sciences Reference Style
\usepackage{graphicx}%
\usepackage{multirow}%
\usepackage{amsmath,amssymb,amsfonts}%
\usepackage{amsthm}%
\usepackage{mathrsfs}%
\usepackage[title]{appendix}%
\usepackage{xcolor}%
\usepackage{textcomp}%
\usepackage{manyfoot}%
\usepackage{booktabs}%
\usepackage{algorithm}%
\usepackage{algorithmicx}%
\usepackage{algpseudocode}%
\usepackage{listings}%

\usepackage{acronym}
\usepackage{adjustbox}
\usepackage{booktabs}
\usepackage[font=small,labelfont=bf]{caption}
\usepackage{colortbl}
\usepackage{dashrule}
\usepackage{fmtcount}
\usepackage{forloop}
\usepackage{stackengine}
\usepackage{subfig}
\usepackage{suffix}
\usepackage{svg}
\usepackage{rotating}
\usepackage{pifont}    % ding
\usepackage{stmaryrd}  % iverson brackets
\usepackage{tabularx}
\usepackage[textsize=scriptsize]{todonotes}
\usepackage{xfrac}
\usepackage{xr}
\usepackage{xspace}
\usepackage{pdflscape}

\presetkeys{todonotes}{inline}{}  % Inline by default

\makeatletter
\newcommand*{\addFileDependency}[1]{% argument=file name and extension
  \typeout{(#1)}
  \@addtofilelist{#1}
  \IfFileExists{#1}{}{\typeout{No file #1.}}
}
\makeatother

\definecolor{ForestGreen}{HTML}{228b22}  % xcolor tends to have option clash

\DeclareRobustCommand{\nbd}{\nobreakdash-} % Usage: "multi{\nbd}scale"

\newcommand{\ndim}[1]{#1{\nbd}D}
\newcommand{\pnorm}[1]{L\ensuremath{_#1}}

\newcommand{\heading}[1]{\noindent \textbf{\small #1.}}

\newcommand{\addfig}[3][htbp]{%
\begin{figure}[#1]
\centering
\input{Figures/#2}
\label{fig:#3}
\end{figure}
}

\WithSuffix\newcommand\addfig*[3][htbp]{%
\begin{figure*}[#1]
\centering
\input{Figures/#2}
\label{fig:#3}
\end{figure*}
}

\newcommand{\addtbl}[3][htbp]{%
\begin{table}[#1]
\footnotesize
\addtolength{\tabcolsep}{-0.2em}
\renewcommand{\arraystretch}{1.2}
\centering
\input{Tables/#2}
\label{tbl:#3}
\end{table}
}

\WithSuffix\newcommand\addtbl*[3][htbp]{%
\begin{table*}[#1]
\footnotesize
\addtolength{\tabcolsep}{-0.2em}
\renewcommand{\arraystretch}{1.2}
\centering
\input{Tables/#2}
\label{tbl:#3}
\end{table*}
}

\def\eg{\emph{e.g}.~} 
\def\ie{\emph{i.e}.~}

\def\wrt{w.r.t.~} 

\def\etal{\emph{et al}.\xspace}

\newcommand{\fig}[1]{Figure~\ref{fig:#1}}
\newcommand{\tbl}[1]{Table~\ref{tbl:#1}}
\newcommand{\sct}[1]{Section~\ref{sec:#1}}

\newcommand{\eq}[1]{(\ref{eq:#1})}

\newcommand{\acromath}[3]{\acrodef{#1}[\(#2\)]{#3}} 
\newcommand{\myac}[1]{\text{\acs{#1}}}  % For inserting acronyms in acronyms or as subscripts!

\DeclareRobustCommand{\asum}{\ensurestackMath{\mathop{\mathchoice%
  {\stackon[-3.8ex]{\displaystyle\sum}{\smash{\rule{.4pt}{4ex}}}}%
  {\stackon[-2.6ex]{\textstyle\sum}{\smash{\rule{.4pt}{2.9ex}}}}%
  {\stackon[-1.9ex]{\scriptstyle\sum}{\smash{\rule{.4pt}{2.2ex}}}}%
  {\stackon[-1.4ex]{\scriptscriptstyle\sum}{\smash{\rule{.4pt}{1.7ex}}}}%
}}}

\DeclareMathOperator*{\mymin}{min}

\newcommand{\minus}{\scalebox{0.5}[0.75]{\ensuremath{-}}}
\newcommand{\inv}{\minus 1}

\newcommand{\shape}[4]{\ensuremath{  % a x b x c x d
#1 \times #2
\ifthenelse {\equal{#3}{}} {} {\times #3}
\ifthenelse {\equal{#4}{}} {} {\times #4}
}}
\newcommand{\sca}[1]{\ensuremath{#1}}  						   % Non-bold
\newcommand{\vct}[1]{\ensuremath{\textbf{\MakeLowercase{#1}}}} % Lowercase/Bold
\newcommand{\mat}[1]{\ensuremath{\textbf{\MakeUppercase{#1}}}} % Uppercase/Bold
\newcommand{\weight}{\lambda}

\newcommand{\Loss}{\mathcal{L}}
\newcommand{\Mask}{\ensuremath{\mathbb{M}}}

\newcommand{\concat}{\ensuremath{\oplus}\xspace}

\newcommand{\brackets}[3]{\ensuremath{\left#1 #2 \right#3}}
\newcommand{\mypar}[1]{\brackets{(}{#1}{)}}
\newcommand{\mybra}[1]{\brackets{\{}{#1}{\}}}

\newcommand{\mymag}[1]{\brackets{|}{#1}{|}}  

\newcommand{\mybil}[1]{\brackets{\langle}{#1}{\rangle}}  % < a >
\newcommand{\myivr}[1]{\brackets{\llbracket}{#1}{\rrbracket}}  % [[ a ]]

\newcommand{\easyfunc}[2]{\ensuremath{#1\!\mypar{#2}}}                 % a(b)
\newcommand{\func}[3]{\ensuremath{ #1 = \easyfunc{#2}{#3} }}         % a = b(c)
\newcommand{\funcdef}[3]{\ensuremath{ \easyfunc{#1}{#2} = #3 }}      % a(b) = c

\acrodef{ar}     [AR]             {Augmented Reality}
\acrodef{araug}  [AR-Aug]         {aspect ratio augmentation}

\acrodef{cnn}    [CNN]            {Convolutional Neural Network}

\acrodef{d25}    [$\delta_{.25}$] {$\delta < 1.25$}
\acrodef{dnn}    [DNN]            {Deep Neural Network}

\acrodef{fscore} [F]              {F-Score}

\acrodef{imu}    [IMU]            {Inertial Measurement Unit}

\acrodef{kbr}    [KBR]            {Kick Back \& Relax}
\acrodef{kbr++}  [KBR++]          {Kick Back \& Relax++}

\acrodef{lidar}  [LiDAR]          {Light Detection and Ranging}

\acrodef{mae}    [MAE]            {Mean Absolute Error}
\acrodef{mde}    [MDE]            {monocular depth estimation}
\acrodef{ml}     [ML]             {Machine Learning}

\acrodef{rank}    [Rank]            {Mean Rank}
\acrodef{rel}    [Rel]            {Absolute Relative Error}
\acrodef{relimp} [$\Delta$]       {Relative Improvement}

\acrodef{sfm}    [SfM]            {Structure-from-Motion}
\acrodef{slam}   [SLAM]           {Simultaneous Localization and Mapping}
\acrodef{sota}   [SotA]           {State-of-the-Art}
\acrodef{ssl}    [SSL]            {self-supervised learning}

\acrodef{vo}     [VO]             {Visual Odometry}

\acrodef{crf}      [NeWCRFs] {NeWCRFs}
\acrodef{ctv}      [CTV]     {CribsTV}

\acrodef{ddad}     [DDAD]    {DDAD}
\acrodef{dii}      [DIODE]   {DIODE Indoors}
\acrodef{dio}      [DIODE]   {DIODE Outdoors}
\acrodef{dpt-beit} [DPT]     {DPT-BEiT}
\acrodef{dpt-vit}  [DPT]     {DPT-ViT}

\acrodef{kb}       [KB]      {Kitti Benchmark}
\acrodef{ke}       [KE]      {Kitti Eigen}
\acrodef{keb}      [Kitti]     {Kitti Eigen{\nbd}Benchmark}
\acrodef{kez}      [KEZ]     {Kitti Eigen{\nbd}Zhou}

\acrodef{mc}       [Mannequin]      {Mannequin Challenge}
\acrodef{midas}    [MiDaS]   {MiDaS}

\acrodef{nyud}     [NYUD]    {NYUD{\nbd}v2}

\acrodef{sintel}   [Sintel]  {Sintel}
\acrodef{ssmde}    [SS-MDE]  {SS{\nbd}MDE}
\acrodef{stv}      [STV]     {SlowTV}
\acrodef{syns}     [SYNS]    {SYNS{\nbd}Patches}

\acrodef{tum}      [TUM]     {TUM{\nbd}RGBD}

\acromath{Cam} {\mat{K}} {}

\acromath{fxy} {\vct{f}_{xy}} {}
\acromath{cxy} {\vct{c}_{xy}} {}

\acromath{Depth} {\mat{D}} {}
\acromath{Depth-t} {\myac{Depth}_{\myac{time}}} {}
\acromath{Disp} {\hat{\ac{Depth}}} {}
\acromath{Disp-t} {\myac{Disp}_{\myac{time}}} {}
\acromath{Disp-tk} {\myac{Disp}_{\myac{tk}}} {}

\acromath{Img} {\mat{I}} {}
\acromath{Img-synth} {\myac{Img}'} {}
\acromath{Img-synth-tk} {\myac{Img-synth}_{\myac{tk}}} {}
\acromath{Img-t} {\myac{Img}_{\myac{time}}} {}
\acromath{Img-tk} {\myac{Img}_{\myac{tk}}} {}

\acromath{Loss-l1}{\sca{\Loss}_{\textit{1}}}{}
\acromath{Loss-ssim}{\sca{\Loss}_{\textit{ssim}}}{}
\acromath{Loss-photo}{\sca{\Loss}_{\textit{ph}}}{}
\acromath{Loss-rec}{\sca{\Loss}_{\textit{rec}}}{}

\acromath{Mask-static}{\mat{\Mask}}{}

\acromath{Net-Depth} {\sca{\Phi}_{\textit{D}}} {}
\acromath{Net-Pose} {\sca{\Phi}_{\textit{P}}} {}

\acromath{offset} {\sca{k}} {}
\acromath{pix} {\vct{p}} {}
\acromath{pix-t} {\myac{pix}_{\myac{time}}} {}
\acromath{pix-synth} {\myac{pix}'} {}
\acromath{pix-synth-tk} {\myac{pix-synth}_{\myac{tk}}} {}
\acromath{Pose} {\hat{\mat{P}}} {}
\acromath{Pose-tk} {\myac{Pose}_{\myac{tk}}} {}

\acromath{time} {\sca{t}} {}
\acromath{tk}{\myac{time}\!+\!\myac{offset}} {}

\acromath{weight-ssim}{\sca{\weight}}{}

\acromath{target}{\sca{y}} {}
\acromath{pred}{\hat{\myac{target}}} {}
\def\codelink{\url{https://github.com/jspenmar/slowtv_monodepth}}
\graphicspath{{Images/}}

\begin{document}

\title[Kick Back \& Relax++]{Kick Back \& Relax++: 
Scaling Beyond Ground-Truth Depth with          %% Simon's suggestion
%Learning to Reconstruct the World  by Watching           %% original title
SlowTV \& CribsTV}


\author*[1]{\fnm{Jaime} \sur{Spencer}}\email{j.spencermartin@surrey.ac.uk}
\author[2]{\fnm{Chris} \sur{Russell}}\email{chris.russell@oii.ox.ac.uk}
\author[1]{\fnm{Simon} \sur{Hadfield}}\email{s.hadfield@surrey.ac.uk}
\author[1]{\fnm{Richard} \sur{Bowden}}\email{r.bowden@surrey.ac.uk}

\affil*[1]{\orgdiv{CVSSP}, \orgname{University of Surrey}}
\affil[2]{\orgname{Oxford Internet Institute}}
% \affil[3]{\orgdiv{Department}, \orgname{Organization}, \orgaddress{\street{Street}, \city{City}, \postcode{610101}, \state{State}, \country{Country}}}


\abstract{
Self-supervised learning is the key to unlocking generic computer vision systems.
By eliminating the reliance on ground-truth annotations, it allows scaling to much larger data quantities. 
Unfortunately, self-supervised monocular depth estimation (SS-MDE) has been limited by the absence of diverse training data. 
Existing datasets have focused exclusively on urban driving in densely populated cities, resulting in models that fail to generalize beyond this domain.

To address these limitations, this paper proposes two novel datasets: SlowTV and CribsTV. 
These are large-scale datasets curated from publicly available YouTube videos, containing a total of 2M training frames.
They offer an incredibly diverse set of environments, ranging from snowy forests to coastal roads, luxury mansions and even underwater coral reefs.
We leverage these datasets to tackle the challenging task of zero-shot generalization, outperforming every existing SS-MDE approach and even some state-of-the-art supervised methods.

The generalization capabilities of our models are further enhanced by a range of components and contributions: 1) learning the camera intrinsics, 2) a stronger augmentation regime targeting aspect ratio changes, 3) support frame randomization, 4) flexible motion estimation, 5) a modern transformer-based architecture.  
We demonstrate the effectiveness of each component in extensive ablation experiments. 
To facilitate the development of future research, we make the datasets, code and pretrained models available to the public at \codelink.
}

\keywords{Monocular Depth Estimation, Self-Supervised Learning, Zero-shot Generalization, Large-scale Dataset Curation}



\maketitle

% ==============================================================================
% ------------------------------------------------------------------------------
% ..............................................................................
% =====================================

% \todo[caption=]{
% \begin{itemize}
%     % \item Add all missing citations
%     \item Arrange figures and tables
%     % \item Reword abstract \& add journal contributions --> DONE
%     \item Reword lit review
%     % \item Reword methodology (networks, losses, camera intrinsics)
%     \item Reword results (ablations, implementations details)
%     \item Incorporate supplementary material?
% \end{itemize}
% }

% =====================================
%
\begin{figure}[htbp]
\centering
\input{Figures/*}
\label{fig:[}
\end{figure}
t]{intro}{intro}
% =====================================

% ------------------------------------------------------------------------------
\section{Introduction} \label{sec:intro}

Reliably reconstructing the \ndim{3} structure of the world is a crucial component in many real-world applications, such as autonomous driving, robotics, camera relocalization or augmented reality. 
While traditional depth estimation algorithms relied on correspondence estimation and triangulation, recent research has shown it is possible to train a neural network to reconstruct a scene from only a single image. 
Despite being an ill-posed task due to the scale ambiguity, \ac{mde} has become of great interest due to its flexibility and applicability to many fields. 

Recent supervised \ac{mde}~\cite{Ranftl2020,Ranftl2021,Bhat2021,Weihao2022} approaches have achieved impressive results, but are limited both by the availability and quality of annotated datasets. 
\acs{lidar} data is expensive to collect and frequently exhibits boundary artifacts due to motion correction. 
Meanwhile, \ac{sfm} is computationally expensive and can produce noisy, incomplete or incorrect reconstructions.

\Ac{ssl} should be able to scale to much larger data quantities, since only monocular or stereo video is required for training. 
These models instead leverage photometric constraints, using the predicted depth and motion to warp adjacent frames and synthesize the target image. 
However, in practice, existing \acs{ssmde} models~\cite{Garg2016,Zhou2017,Godard2019,Lyu2021} rely exclusively on automotive datasets~\cite{Geiger2013,Cordts2016,Guizilini2020}.
This lack of variety significantly impacts their generalization capabilities and results in failures when applied to natural or indoor scenes. 
Moreover, despite being fully convolutional, these models struggle to generalize to different image sizes. 

We argue that the lack of diversity is due to the challenges of data collection, with new datasets aiming to provide high-quality ground-truth annotations that can be used for testing~\cite{Adams2016,Koch2018}.
Whist this is important to accurately evaluate the performance of models, it also places strong limitations on the achievable scale for the training splits. 
In this paper, we instead focus on creating datasets that specifically target self-supervised learning, exploiting the fact that no ground-truth annotations are required. 
Combined with our additional contributions, we train self-supervised models capable of zero-shot generalization beyond the automotive domain. 
Our models significantly outperform all existing \acs{ssmde} approaches and can even match or outperform \ac{sota} \emph{supervised} techniques. 

A preliminary version of this work~\cite{Spencer2023c} was published at the International Conference on Computer Vision. 
This paper introduced the \textbf{\acl{stv}} dataset, composed of 1.7M frames from 40 curated YouTube videos. 
These videos featured a wide diversity of settings, including seasonal hiking, scenic driving and scuba diving. 
\acl{stv} provided our models with the general foundation for natural scenes, such as forests, mountainous terrains or deserts. 
Our dataset was combined with \acl{mc}~\cite{Li2020} and Kitti~\cite{Geiger2013}, which targeted indoor scenes with humans and urban driving. 

Our preliminary work introduced several contributions that further maximized zero-shot performance, whilst not increasing the complexity and computational requirements of the model. 
For instance, predicting the camera intrinsics prevented performance drops resulting from training with inaccurate intrinsics estimated via \ac{sfm}. 
Meanwhile, the \ac{araug} diversified the distribution of image shapes seen during training and facilitated transfer across datasets.
% Finally, we demonstrated that randomizing the gap between supplementary frames and removing the forward-only motion constraint further improved the model's flexibility. 

This paper presents several significant extensions to \ac{kbr}~\cite{Spencer2023c} and thus introduces \acs{kbr++}.
Despite performing on par with several supervised \ac{sota} models, there was still a gap when evaluating on indoor datasets due to the exclusive reliance on \acl{mc} as a source of indoor data.
Following the design philosophy of our original work, we introduce the \textbf{\acl{ctv}} dataset. 
This is an extension to \acl{stv} consisting of 330k images from curated YouTube real estate virtual tours. 
As such, this new dataset focuses on bedrooms, living rooms and kitchens and is complemented by gardens, swimming pools and aerial outdoor shots. 
This further reduces the gap between supervision and self-supervision in indoor settings. 

We complement this novel dataset with additional augmentation strategy experiments.
Since its inception~\cite{Garg2016,Zhou2017,Godard2017}, \acs{ssmde} has restricted itself to simple augmentations, such as color jittering and horizontal flipping. 
However, recent contrastive \ac{ssl}~\cite{Chen2020,Caron2021,Bao2022} research has shown the benefits of more aggressive augmentation schemes. 
We demonstrate this is also the case in \acs{ssmde} and incorporate RandAugment~\cite{Cubuk2020} and CutOut~\cite{Devries2017} into the pipeline.
Finally, we modernize the depth network architecture with the transformer-based backbones from DPT~\cite{Ranftl2021} and perform several new ablation experiments that give further insight into the performance of our models. 
Our updated models outperform all (self-)supervised approaches, except \acl{dpt-beit}, despite not requiring any ground-truth annotations.

The contributions of our works can be summarized as:
% =====================================
\begin{enumerate}
    \item We introduce a novel \acl{ssmde} dataset of \acl{stv} YouTube videos and complement it with \acl{ctv}, resulting in a total of 2M training images.
    This dataset features an incredibly diverse set of environments, including worldwide seasonal hiking, scenic driving, scuba diving and real estate tours. 

    \item We leverage \acl{stv} and \acl{ctv} to train zero-shot models that generalize across multiple datasets.
    We additionally apply these models to the task of map-free relocalization, demonstrating their applicability to real-world settings. 

    \item We introduce a range of contributions and best-practices that further maximize generalization. 
    This includes: camera intrinsics learning, an aspect ratio augmentation, stronger photometric augmentations, support frame randomization, flexible motion estimation and a modernized depth network architecture.
    We demonstrate the effectiveness of these contributions in detailed ablation experiments.

    \item We close the performance gap between supervision and self-supervision, greatly furthering the \ac{sota} in \acs{ssmde}.
    We share these developments with the community, making the datasets, pretrained models and code available to the public. 
\end{enumerate}
% =====================================

%------------------------------------------------------------------------
\section{Related Work} \label{sec:lit}
Instead of using ground-truth depth annotations from \acs{lidar} or RGB-D sensors, self-supervised monocular depth estimation relies exclusively on photometric consistency constraints. 
The seminal approach by Garg~\etal~\cite{Garg2016} combined the known baseline between stereo pairs with the predicted depth to obtain correspondences and perform view synthesis. 
Monodepth~\cite{Godard2017} complemented this with the virtual stereo consistency loss.
Performance was further improved by introducing differentiable bilinear interpolation~\cite{Jaderberg2015} and a reconstruction loss based on SSIM~\cite{Wang2004}. 
3Net~\cite{Poggi2018} extended the virtual stereo consistency into a trinocular setting.

To extend this formulation to the monocular domain, it is necessary to replace the fixed stereo baseline with a network to predict the relative pose between frames. 
This was first proposed by SfM-Learner~\cite{Zhou2017} and extended by DDVO~\cite{Wang2018}, which introduced a differentiable DSO module~\cite{Engel2018}.
Purely monocular approaches are sensitve to dynamic objects, as their additional motion is not accounted-for by the relative pose estimation. 
This results in incorrect correspondences, which further lead to inaccurate depth predictions. 
Therefore, future research aimed to minimize this impact via predictive masking~\cite{Zhou2017}, uncertainty estimation~\cite{Klodt2018,Yang2020a,Poggi2020}, optical flow~\cite{Yin2018,Ranjan2019,Luo2020} or motion masks~\cite{Gordon2019,Casser2019,Dai2020}.

Several works have instead focused on improving the robustness of the photometric loss. 
One notable example is Monodepth2~\cite{Godard2019}, which introduced the minimum reconstruction loss and static-pixel automasking. 
FeatDepth~\cite{Zhan2018} applied the same view synthesis to dense feature descriptors, which should be invariant to viewpoint and illumination conditions. 
DeFeat-Net~\cite{Spencer2020} learned the feature descriptors simultaneously, while Shu~\etal~\cite{Shu2020} used intermediate autoencoder representations. 
Others complemented the photometric loss with semantic segmentation~\cite{Chen2019,Guizilini2020b,Jung2021} or geometric constraints~\cite{Mahjourian2018,Bian2019,Watson2021}.
Finally, is is also common to introduce proxy depth label regression obtained from \acs{slam}~\cite{Klodt2018,Rui2018}, synthetic data~\cite{Luo2018} or hand-crafted disparity~\cite{Tosi2019,Watson2019}.

The encoder network architecture has been improved by introducing \ndim{3} (un)packing blocks, positional encoding~\cite{Bello2021}, transformers~\cite{Zhao2022} or high-resolution networks~\cite{Zhou2021}. 
Updated decoders have focused on sub-pixel convolutions~\cite{Pillai2019}, self-attention~\cite{Johnston2020,Zhou2021,Yan2021} and progressive skip connections~\cite{Lyu2021}. 
Akin to supervised \ac{mde} developments~\cite{Bhat2021,Bhat2022}, Johnston~\etal~\cite{Johnston2020} and Bello~\etal~\cite{Bello2020,Bello2021} obtained improvements by representing depth as a discrete volume.

So far, these contributions have only been tested on automotive datasets, such as Kitti~\cite{Geiger2013}, CityScapes~\cite{Mayer2016} or DDAD~\cite{Guizilini2020}.
Recent benchmarks and challenges~\cite{Spencer2022,Spencer2023,Spencer2023b} have shown that these models fail to generalize beyond this restricted training domain.
Meanwhile, recent supervised models~\cite{Ranftl2020,Ranftl2021,Spencer2023b} have leveraged collections of datasets to improve zero-shot generalization. 
In this paper, we aim to close the gap between supervised and self-supervised \ac{mde} in the challenging task of zero-shot generalization. 
This is achieved by greatly increasing the diversity and scale of the training data by leveraging unlabeled videos from YouTube, without requiring manual annotation or expensive pre-processing.

% =====================================
%
\begin{figure}[htbp]
\centering
\input{Figures/*}
\label{fig:[}
\end{figure}
t]{datasets}{data:viz}
% =====================================

% =====================================
%
\begin{figure}[htbp]
\centering
\input{Figures/*}
\label{fig:[}
\end{figure}
t]{supp_slowtv_map}{data:map}
% =====================================

% ------------------------------------------------------------------------------
\section{Datasets} \label{sec:data}

The proposed \textbf{\acl{stv}} and \textbf{\acl{ctv}} datasets consist of 45 videos curated from YouTube, totaling more than 140 hours of content and 2 million training images.  
As shown in \tbl{data:overview}, this is an order of magnitude more data than any commonly used \acs{ssmde} dataset.
\acl{stv} contains three main outdoor categories (hiking, driving and scuba diving), while \acl{ctv} focuses exclusively on real estate properties.
When combined, these datasets provide an incredibly diverse set of training scenes for our models, allowing us to tackle the challenging task of zero-shot generalization.

% =====================================
%
\begin{table}[htbp]
\footnotesize
\addtolength{\tabcolsep}{-0.2em}
\renewcommand{\arraystretch}{1.2}
\centering
\input{Tables/*}
\label{tbl:[}
\end{table}
t]{datasets}{data:overview}
% =====================================

Hiking videos target natural scenes, such as forest, mountains, deserts or fields, which are non-existent in current datasets.
Our driving split seeks to complement existing automotive datasets, which tend to focus on urban driving in densely populated cities~\cite{Geiger2013,Cordts2016,Huang2019,Chang2019,Guizilini2020,Yu2020,Ceasar2020}. 
The proposed split instead features videos from scenic routes, traversing forest, mountainous or coastal roads with sparse traffic. 
We also feature a variety of weather and seasonal conditions. 
Underwater scuba diving represents yet another previously unexplored domain, which further increases data diversity. 
Finally, real estate properties are a natural counterpart to the previous outdoor data. 
They also complement the \acl{mc}~\cite{Li2020}, which primarily focuses on human beings in indoor settings, rather than the indoor scenes themselves. 

The proposed videos were collected from a diverse set of locations and conditions, as illustrated in \fig{data:map}. 
This includes the USA, Canada, the Balkans, Eastern Europe, Indonesia and Hawaii, and conditions such as rain, snow, autumn and summer. 
Since \acl{ctv} contains a large number of individual properties, it is challenging to obtain accurate information about each of their locations. 
However, they are predominantly located in the USA. 
\fig{data:viz} shows sample frames from each of the available categories, illustrating the dataset's incredible diversity.

Videos were downloaded at HD resolution (\shape{720}{1280}{}{}) and extracted at 10 FPS to reduce storage, while still providing smooth motion and large overlap between adjacent frames. 
In the case of \acl{stv}, only 100 consecutive frames out of every 250 were retrained.
This reduces the self-similarity between training samples and keeps the dataset size tractable. 
The final \acl{stv} contains a total of 1.7M images, composed of 1.1M natural, 400k driving and 180k underwater.

Since we target \acs{ssmde}, the only annotations required are the camera intrinsic parameters, which can be estimated using COLMAP~\cite{Schoenberger2016}.
However, as discussed in \sct{meth:cam}, it is possible to let the network jointly optimize camera parameters alongside depth and motion.
We find that this is more robust and improves performance compared to training with potentially inaccurate COLMAP intrinsics.

\acl{ctv} instead consists of highly-produced cinematic house tours. 
In practice, this means they include cuts between shots of different viewpoints or each room. 
Some of these shots may also be unsuitable for \acs{ssmde}, containing zooming/focusing/blurring effects or static shots.
As such, it was first necessary to split each video into its individual scenes using an off-the-shelf scene detector~\cite{Castellano2014}.  
We then performed a quick manual check to filter out potentially invalid scenes based on their first frame.
The final dataset contains 330k training frames. 

\acl{ctv} also presents additional challenges when estimating the camera intrinsics. 
Due to the short shot duration (on average 5 seconds) and lack of overlap between scenes, COLMAP is unable to produce any reconstructions. 
This again motivates the need for a more flexible depth estimation framework, capable of estimating camera intrinsics.
This reduces the complexity of dataset collection and allows us to train with much larger quantities of diverse data.

% ------------------------------------------------------------------------------
\section{Methodology} \label{sec:meth}

Monocular depth estimation aims to reconstruct the \ndim{3} structure of the scene using only a single \ndim{2} image projection. 
However, additional support frames are required in order to synthesize the target view and compute the photometric reconstruction losses that drive optimization.
In the case where only stereo pairs are used~\cite{Garg2016,Godard2017}, the predicted depth is combined with the known stereo baseline to perform the view synthesis. 
However, if only monocular video is available, such as YouTube videos from \acl{stv} or \acl{ctv}, is becomes necessary to incorporate an additional pose network to estimate the relative motion between adjacent frames~\cite{Zhou2017}.
A key difference between these forms of supervision is that stereo approaches can estimate metric depth, while monocular approaches are only accurate up to unknown scale and shift factors.

\heading{Depth}
The the depth estimation network~\acs{Net-Depth} can be formalized as 
% =====================================
$
    \func{\acs{Disp-t}}{\acs{Net-Depth}}{\acs{Img-t}},    
$
% =====================================
where \acs{Disp-t} is the predicted sigmoid disparity and \acs{Img-t} is the target image.
Note that the disparity map must be inverted into a depth map and appropriately scaled in order to warp the support images.

\heading{Pose}
Similarly, the pose estimation network is represented as 
% =====================================
$
    \func{\acs{Pose-tk}}{\acs{Net-Pose}}{\acs{Img-t} \concat \acs{Img-tk} },    
$
% =====================================
where \concat is channel-wise concatenation, \acs{Img-tk} is the support frame at time offset $\acs{offset} \in \mybra{\minus1, +1}$ and \acs{Pose-tk} is the predicted motion as a translation and axis-angle rotation.

% ..............................................................................
\subsection{Losses} \label{sec:meth:losses}
Supervised approaches such as MiDaS~\cite{Ranftl2020}, DPT~\cite{Ranftl2021} or NewCRFs~\cite{Weihao2022} require ground-truth depth annotations, in the form of LiDAR, depth cameras, \acs{sfm} reconstructions or stereo disparity estimation. 
\acs{ssmde}~\cite{Garg2016,Godard2019,Lyu2021,Zhou2021} instead relies on the photometric consistency across the target and support frames. 

Using the predicted depth~\acs{Depth-t} and motion~\acs{Pose-tk}, pixel-wise correspondences between these images can be obtained as 
% =====================================
\begin{equation} \label{eq:reprojection}
    \acs{pix-synth-tk}
    =
    \acs{Cam}
    \acs{Pose-tk}
    \easyfunc{\acs{Depth-t}}{\acs{pix-t}}
    \acs{Cam}^{\inv} 
    \acs{pix-t}
    ,
\end{equation}
% =====================================
where \acs{Cam} represents a camera's intrinsics parameters, \acs{pix-t} are the \ndim{2} pixel coordinates in the target image and \acs{pix-synth-tk} are its \ndim{2} reprojected coordinates onto the corresponding support frame. 
The synthesized support frame is then obtained via
% =====================================
$
    \acs{Img-synth-tk}
    =
    \acs{Img-tk} \mybil{ \acs{pix-synth-tk} }
    ,
$
% =====================================
where $\mybil{\cdot}$ represents differentiable bilinear interpolation~\cite{Jaderberg2015}.

The reconstruction loss is then given by the weighed combination of SSIM+\pnorm{1}~\cite{Godard2017}, defined as 
% =====================================
\begin{equation}
    \funcdef
        { \acs{Loss-photo} }
        { \acs{Img}, \acs{Img-synth} }
        { 
            \acs{weight-ssim}
            \frac{ 1 \minus \easyfunc{ \acs{Loss-ssim} }{ \acs{Img}, \acs{Img-synth} } }{2}
            + 
            \mypar{ 1 \minus \acs{weight-ssim} }
            \easyfunc{ \acs{Loss-l1} }{ \acs{Img}, \acs{Img-synth} }
        } 
    .
\end{equation}
% =====================================
As is common, the loss balancing weight is set to $\acs{weight-ssim} = 0.85$.

It is well know that purely-monocular approaches~\cite{Zhou2017} are sensitive to artifacts caused by dynamic objects.
This is due to the additional motion of the object being unaccounted for in the correspondence estimation procedure from \eq{reprojection}.
Recent research~\cite{Gordon2019,Casser2019,Dai2020} aimed to solve these challenges using motion masks or semantic segmentation maps. 
Whilst effective, the additional annotations and labeling required makes these contributions unsuitable for the proposed framework.
Instead we opt for the simple, yet effective, contributions from Monodepth2~\cite{Godard2019}.
This includes the minimum reconstruction loss and static-pixel automasking. 

The minimum reconstruction loss reduces the impact of occlusions by assuming only support frames contains a correct correspondence. 
This frame is obtained by finding the minimum pixel-wise error across all support frames, defined as 
% =====================================
\begin{equation}
    \acs{Loss-rec}
    =
    \asum_{\myac{pix}} 
    \mymin_{\myac{offset}} 
    \easyfunc
        { \acs{Loss-photo} }
        { \acs{Img-t}, \acs{Img-synth-tk} }
    ,
\end{equation}
where $\asum$ indicates averaging over a set.
% =====================================

Automasking instead reduces the effect of static frames and objects moving at the same speed as the camera.
These objects remain static across frames, giving the impression of an infinite depth.
Automasking simply removes pixels from the loss where the original \emph{non-warped} support frame has a lower reconstruction error that the synthesized view. 
This is computed as 
% =====================================
\begin{equation}
    \acs{Mask-static}
    =
    \myivr{ 
        \mymin_{\myac{offset}} 
        \easyfunc{ \acs{Loss-photo} }{ \acs{Img-t}, \acs{Img-synth-tk} }
        <
        \mymin_{\myac{offset}} 
        \easyfunc{ \acs{Loss-photo} }{ \acs{Img-t}, \acs{Img-tk} }
    }
    ,
\end{equation}
% =====================================
where $\myivr{\cdot}$ represents the Iverson brackets.

Whilst being simple to implement, \fig{meth:motion} demonstrates the effectiveness of incorporating these contributions.
Finally, we complement the reconstruction loss with the common edge-aware smoothness regularization~\cite{Godard2017}.
These networks and losses constitute the core baseline required to train the desired zero-shot depth estimation models. 
The following sections describe additional contributions that help to further maximize performance and generalization capabilities.

% =====================================
%
\begin{figure}[htbp]
\centering
\input{Figures/*}
\label{fig:[}
\end{figure}
t]{viz_motion}{meth:motion}
% =====================================

% ..............................................................................
\subsection{Learning Camera Intrinsics} \label{sec:meth:cam}
Many datasets provide accurately calibrated camera intrinsic parameters. 
Unfortunately, in crowd-sourced~\cite{Arnold2022}, photo-tourism~\cite{Li2018} or internet-curated datasets (such as \acl{stv} or \acl{ctv}) these parameters are not freely available. 
Instead, it is common practice to rely on \ac{sfm} reconstructions obtained from COLMAP~\cite{Schoenberger2016}. 
Unfortunately, these reconstructions may be incorrect, incomplete or sometimes impossible to obtain.
As such, especially in the case of internet-curated datasets that may continuously change or scale up in size, it would be extremely beneficial to omit these pre-processing requirements. 

We take inspiration from~\cite{Gordon2019,Chen2019b} learn depth, pose and camera intrinsics simultaneously. 
To achieve this, two additional branches are incorporated into the pose estimation network as 
% =====================================
\begin{equation}
    \func
        {\acs{Pose-tk}, \acs{fxy}, \acs{cxy}}
        {\acs{Net-Pose}}
        {\acs{Img-t} \concat \acs{Img-tk} },    
\end{equation}
% =====================================
where \acs{fxy} and \acs{cxy} represent the focal lengths and principal point, respectively. 
These parameters are predicted as normalized and scaled accordingly based on the input image size. 
The branch predicting the focal lengths uses a softplus activation to ensure a positive output. 
The principal point instead uses sigmoid, under the assumption that it will lie within the image plane. 

Incorporating these decoders results in a negligible 2MParam increase in the pose estimation network, with no additional computation required for the losses. 
Instead, we simply modify \eq{reprojection} to use the predicted intrinsics instead of the ground-truth ones. 
Despite this, our ablations in \sct{res:abl} show that this increases performance compared to training with intrinsics estimated by COLMAP. 

% ..............................................................................
\subsection{Augmentation Strategies} \label{sec:meth:aug}
Existing research~\cite{Zhang2017,Muller2021,Cubuk2020} has shown the importance of incorporating more sophisticated augmentation strategies. 
This is especially the case in \ac{ssl}, which relies exclusively on the the diversity of the available data. 
However, existing \ac{ssmde} approaches use only traditional augmentations such as color jittering and horizontal flipping. 
This section describes the additional augmentations incorporated into our training regime to further boost the generalization capabilities of our final models.

% =====================================
%
\begin{figure}[htbp]
\centering
\input{Figures/*}
\label{fig:[}
\end{figure}
t]{distort}{meth:distort}
% =====================================

% ,,,,,,,,,,,,,,,,,,,,,,,,,,,,,,,,,,,,,,,,,,,,,,,,,,,,,,,,,,,,,,,,,,,,,,,,,,,,,,
\subsubsection{Aspect Ratio} \label{sec:meth:araug}
Dense predictions networks, such as the depth network used in this paper, can process images of arbitrary shape.
However, when trained only on a single dataset with a fixed image size, it is common for them to overfit to this size, resulting in poor performance on out-of-dataset examples. 
An example of this effect can be seen in \fig{meth:distort}, where first resizing the image to match the training aspect ratio results in better performance, despite introducing stretching or squashing artifacts. 

We overcome this by introducing an augmentation that randomizes the image sizes and aspect ratios seen by the network during training. 
The proposed \acl{araug} augmentation (\acs{araug}) consists of two components: cropping and resizing. 
The first stage uniformly samples from a set of predefined common aspect ratios\footnote{%
Portrait: 6:13, 9:16, 3:5, 2:3, 4:5, 1:1. 
Landscape: 5:4, 4:3, 3:2, 14:9, 5:3, 16:9, 2:1, 24:10, 33:10, 18:5.
}.
A random crop is generated using this aspect ratio, covering 50-100\% of original image height or width. 
The resizing stage ensures that the final crop has roughly the same number of pixels as the original input image. 
\fig{meth:araug} shows training samples obtained using this procedure.

This augmentation is applied at the mini-batch level to ensure all images are the same shape. 
If using ground-truth intrinsics, these are rescaled accordingly. 
\ac{araug} has the effect of drastically increasing the diversity of shapes and sizes seen by the network and prevents overfitting to a single shape. 

% =====================================
%
\begin{figure}[htbp]
\centering
\input{Figures/*}
\label{fig:[}
\end{figure}
t]{araug}{meth:araug}
% =====================================

% ,,,,,,,,,,,,,,,,,,,,,,,,,,,,,,,,,,,,,,,,,,,,,,,,,,,,,,,,,,,,,,,,,,,,,,,,,,,,,,
\subsubsection{RandAugment} \label{sec:meth:randaug}
% Existing \ac{mde} models use only simple augmentation strategies, such as color jittering and horizontal flipping.
% However, significant effort has been put into developing new augmentations that can further improve performance~\cite{?,?,?}. 
% This is especially crucial in self-supervised learning, which relies exclusively on the diversity of the data to learn more generic representations. 

We additionally proposed to complement color jittering with RandAugment~\cite{Cubuk2020}.
This strategy sequentially applies a random combination of photometric and geometric augmentations. 
Since \ac{mde} requires accurate re-projections across a sequence of images we remove the geometric augmentations (\eg translate, rotate and shear) and focus purely on photometric ones. 
The set of possible augmentations is thus reduced to: identity (\ie no augmentation), auto-contrast, equalization, sharpness, brightness, color and contrast.
At each training iteration, a random subset of three augmentations is chosen and applied to both the target and support frames. 
Sample augmented images using this strategy can be found in \fig{meth:aug}.

% =====================================
%
\begin{figure}[htbp]
\centering
\input{Figures/*}
\label{fig:[}
\end{figure}
t]{augs}{meth:aug}
% =====================================

% ,,,,,,,,,,,,,,,,,,,,,,,,,,,,,,,,,,,,,,,,,,,,,,,,,,,,,,,,,,,,,,,,,,,,,,,,,,,,,,
\subsubsection{CutOut} \label{sec:meth:cutout}
Inspired by the recent success of transformer token-masking augmentations~\cite{Devlin2018,Bao2022,He2022}, we additionally propose to re-introduce CutOut augmentations~\cite{Devries2017}.
While CutOut was originally used to boost holistic tasks like classification, it can also be applied to dense prediction tasks. 
In this case, the objective is to teach the network to predict the depth for a missing region in the image, based only on the context surrounding it. 
As such, these models should learn to incorporate additional context cues and be more robust to test-time artifacts such as reflections or highlights.

To further increase the variability of the augmentations, we implement various fill modes for the masked-out regions: white, black, grayscale, RGB and random.
A different fill mode is randomly selected at each training iteration. 
\fig{meth:aug} shows examples of applying this augmentation and the robustness of the model to it.

% ------------------------------------------------------------------------------
\section{Results} \label{sec:res}
We carry out extensive evaluations to demonstrate the effectiveness of the techniques and datasets proposed in this paper. 
This includes the zero-shot experiments for the final models, as well as detailed ablations on each proposed component. 
We additionally evaluated our models in the challenging task of map-free relocalization~\cite{Arnold2022} and the MDEC-2 challenge~\cite{Spencer2023,Spencer2023b}.

Since both \acs{kbr} and \acs{kbr++} were trained exclusively on monocular data, it is necessary to first align the predictions to the ground-truth metric scale. 
This alignment is obtained using the least-squares procedure proposed by MiDaS~\cite{Ranftl2021} and is applied equally to every baseline.

% ====================================
%
\begin{table}[htbp]
\footnotesize
\addtolength{\tabcolsep}{-0.2em}
\renewcommand{\arraystretch}{1.2}
\centering
\input{Tables/*}
\label{tbl:[}
\end{table}
t]{resources}{res:resources}
% ====================================

% ..............................................................................
\subsection{Baselines} \label{sec:res:baselines}
The \ac{sota} self-supervised models were obtained from the Monodepth Benchmark~\cite{Spencer2022}, which use a pretrained ConvNeXt{\nbd}B backbone. 
These models were trained exclusively on the Kitti dataset~\cite{Eigen2015,Zhou2017} and are therefore also zero-shot on all other datasets.

We also compare our frameworks to current \ac{sota} \emph{supervised} models, which require accurate ground-truth annotations to train.
MiDaS~\cite{Ranftl2020} and DPT~\cite{Ranftl2021} were trained on a collection of 10/12 supervised datasets that do not overlap with our testing set (unless otherwise specified). 
As such, these models are also evaluated in the challenging zero-shot setting. 
We use the pretrained models provided in the PyTorch Hub. 

NewCRFs~\cite{Weihao2022} instead provides separate outdoor/indoor models trained on Kitti and NYUD-v2 respectively. 
We evaluate the corresponding model in a zero-shot manner depending on the dataset category. 
Even though NewCRFs~\cite{Weihao2022} should be capable of predicting metric depth, we apply the least-squares alignment procedue to ensure that all results are comparable. 

% ..............................................................................
\subsection{Implementation Details} \label{sec:res:details}
The proposed models were implemented in PyTorch~\cite{Paszke2019} and based on the Monodepth Benchmark~\cite{Spencer2022}.
The original \acs{kbr}~\cite{Spencer2023c} used a ConvNeXt{\nbd}B backbone~\cite{Liu2022,Wightman2019} and a DispNet decoder~\cite{Mayer2016,Godard2017}. 
The pose network instead used ConvNeXt{\nbd}T for efficiency. 
As such, these models are comparable to the \ac{ssl} baselines~\cite{Spencer2022}.

\tbl{res:resources} shows a comparison between the computational complexity of the proposed models and the supervised \ac{sota}.
As seen, these models use larger transformer-based backbones~\cite{Dosovitskiy2021,Liu2021b,Bao2022}.
In order to make our results more comparable, \acs{kbr++} incorporates the same architecture used by \acs{dpt-beit}~\cite{Ranftl2021,Bao2022}.
Our ablation experiments in \sct{res:abl} show the impact of different backbone architectures.

In our experiments and ablations, each model is trained using three different random seeds and we report average performance. 
This improves the reliability of the results and reduces the impact of non-determinism.
We make the datasets, pretrained models and training code available at \codelink. 

The final \acs{kbr++} models were trained on a combination of \acl{stv} (1.7M), \acl{ctv} (330k), \acl{mc} (115k) and \acl{keb} (71k). 
To make the duration of each epoch tractable and balance the contribution of each dataset, we fix the number of images per epoch to 30k, 15k, 15k and 15k, respectively. 
The subset sampled from each dataset varies with each epoch to ensure a high data diversity.  

The models were trained for 60 epochs using AdamW~\cite{Loshchilov2017} with weight decay $10^{-3}$ and a base learning rate of $10^{-4}$, decreased by a factor of 10 for the final 20 epochs.
Empirically, we found that linearly warming up the learning rate for the first few epochs stabilized learning and prevented model collapse. 
When training with DPT backbones, finetuning the pretrained encoders at a lower learning rate was also found to be beneficial.
We use a batch size of 4 and train the models on a single NVIDIA GeForce RTX 3090.

\acl{stv}, \acl{ctv} and \acl{mc} use a base image size of \shape{384}{640}{}{}, while Kitti uses \shape{192}{640}{}{}.
We apply horizontal flipping and color jittering, along with the proposed RandAugment~\cite{Cubuk2020} and CutOut~\cite{Devries2017} augmentations, each with 30\% probability. 
\ac{araug} is applied with 70\% probability, sampling from 16 predefined aspect ratios previously described.

Since existing models are trained exclusively on automotive data, most of the motion occurs in a straight-line and forward-facing direction. 
It is therefore common practice to force the network to always make a forward-motion prediction by reversing the target and support frame if required. 
Handheld videos, while still primarily featuring forward motion, also exhibit more complex motion patterns.
As such, removing the forward motion constraint results in a more flexible model that improves performance. 

Similarly, existing models are trained with a fixed set of support frames---usually previous and next.
Since \acl{stv} and \acl{mc} are mostly composed of handheld videos, the change from frame-to-frame is greatly reduced. 
We make the model more robust to different motion scales and appearance changes by randomizing the separation between target and support frames. 
In general, we sample such that handheld videos use a wider time-gap between frames, while automotive has a small time-gap to ensure there is significant overlap between frames. 
As shown later, this leads to further improvements and greater flexibility. 

% ..............................................................................
\subsection{Evaluation Metrics} \label{sec:res:metrics}
We follow the original evaluation procedure outlined in~\cite{Spencer2023c} and report the following metrics per dataset:

\heading{\acs{rel}} 
Absolute relative error (\%) between target \acs{target} and prediction \acs{pred} as 
$
    \text{\acs{rel}} = \asum \mymag{\acs{target} \minus \acs{pred}} / \acs{target}.
$

\heading{Delta}
Prediction threshold accuracy (\%) as \\
$
    \acs{d25} = \asum \mypar{ \max \mypar{ \acs{pred}/\acs{target},\ \acs{target}/\acs{pred} } < 1.25 }.
$

\heading{\acs{fscore}}
Pointcloud reconstruction F-Score~\cite{Ornek2022} (\%) as 
$
    \text{\acs{fscore}} = \mypar{2 P R}/\mypar{P + R},
$
where $P$ and $R$ are the Precision and Accuracy of the \ndim{3} reconstruction with a correctness threshold of 10cm.

\vspace{0.2cm}
We additionally compute multi-task metrics to summarize the performance across all datasets: 

\heading{\acs{rank}}
Average ordinal ranking order across all metrics as 
$
    \text{\acs{rank}} = \asum_{m} r_{m},
$
where $m$ represents each available metric and $r$ is the ordinal rank. 

\heading{Improvement}
Average relative increase or decrease in performance (\%) across all metrics as 
$
    \acs{relimp} = \asum_{m} (\minus 1)^{l_m} (M_m - M^0_m) / M^0_m,
$
where $l_m = 1$ if a lower value is better, $M_m$ is the performance for a given metric and $M^0_m$ is the baseline's performance.

% ..............................................................................
\subsection{Ablation} \label{sec:res:abl}
To demonstrate the effectiveness of each proposed component, we carry out a series of ablation studies. 
These experiments generally use a more efficient architecture (ConvNeXt-Tiny) and a smaller training dataset.
% We also opt for a leave-one-out strategy, whereby a single component is removed per-experiment from the full model. 

\heading{Learning \acs{Cam}}
\tbl{res:abl:learnk} shows the benefits of learning the camera intrinsics, as outlined in \sct{meth:cam}. 
We train models on either Kitti or \acl{mc} and test them on the same dataset. 
If the dataset provides accurately calibrated intrinsics (Kitti), this procedure provides comparable performance. 
However, in the case where these were estimated by COLMAP~\cite{Schoenberger2016}, learning \acs{Cam} results in a slight performance boost. 
% As we show in the following experiment, this is more apparent when training/evaluating on multiple datasets. 
This highlights the flexibility of this contribution, which requires only a negligible increase of 2MParams in the pose network, yet allows us to train without ground-truth intrinsics. 
This further simplifies the process of data collection and results in a framework the requires only uncalibrated monocular video to train. 

% =====================================
%
\begin{table}[htbp]
\footnotesize
\addtolength{\tabcolsep}{-0.2em}
\renewcommand{\arraystretch}{1.2}
\centering
\input{Tables/*}
\label{tbl:[}
\end{table}
t]{supp_learnk}{res:abl:learnk}
% =====================================

% =====================================
\begin{landscape}
\thispagestyle{empty}
\begin{table}[htbp]
\footnotesize
\addtolength{\tabcolsep}{-0.2em}
\renewcommand{\arraystretch}{1.2}
\centering
\input{Tables/*}
\label{tbl:[}
\end{table}
t]{res_abl}{res:abl}
\end{landscape}
% =====================================

\heading{\acs{kbr}}
\tbl{res:abl} (1\textsuperscript{st} block) shows results when removing each component proposed by the original \acs{kbr}~\cite{Spencer2023c}.
\emph{Fwd \acs{Pose}} represents a network forced to always make a forward-motion prediction. 
\emph{$\acs{offset} = \pm1$} uses a fixed set of support frames, instead of the randomization proposed in \sct{res:details}.
\emph{Fixed \acs{Cam}} removes the learned camera intrinsics, while \emph{No \acs{araug}} removes the proposed aspect ratio augmentation. 
As expected, the model with the full set of contributions performs best, while the model without any contributions is worse by 7.6\%.
It is interesting to note that both learning the intrinsics and \ac{araug} provide the biggest performance boost. 
Furthermore, it is worth remembering that none of the components (except learning \acs{Cam}) results in a increase in model complexity. 
However, when combined together, they significantly improve the zero-shot generalization capabilities of our models.

\heading{Network Architecture}
To make our models more comparable with the supervised \ac{sota}, we modify our architecture to match the one from DPT~\cite{Ranftl2021}.
These results are shown in \tbl{res:abl} (2\textsuperscript{nd} block).
All models start from an encoder pretrained on ImageNet. 
Interestingly, we find that most transformer-based architectures do not significantly improve upon the baseline (ConvNeXt-B~\cite{Liu2022}), which is much more efficient. 
However, the largest version of BEiT~\cite{Bao2022} provides the best performance.

\heading{Augmentations}
\tbl{res:abl} (3\textsuperscript{rd} block) shows the results of incorporating the more advanced augmentation strategies from \sct{meth:aug}.
All variants (except \emph{No Aug}) additionally include horizontal flipping. 
As shown, the default color jittering augmentation used by most existing models is slightly better than using no augmentations.
However, both CutOut and RandAugment provide larger improvements.
Furthermore, the final row (All) demonstrates that these improvements are cumulative and that the model benefits from combining multiple augmentation strategies. 

\heading{Datasets}
The final ablation experiment explores the effect of removing or adding each training dataset, shown in \tbl{res:abl} (4\textsuperscript{th} block).
As expected, removing each dataset results in a significant drop in performance. 
Whilst \acl{stv} seems to impact performance the least, we believe this is due to the lack of natural data within our evaluation set. 
This is supported by the fact that \acl{stv} has the most impact on \acl{syns}, which is the only dataset with natural scenes. 
Finally, incorporating the \acl{ctv} dataset proposed in this paper slightly increases the overall performance, especially in indoor scenes.
It is worth remembering this variant also includes \acl{mc}, which results in a less drastic improvement.
Furthermore, training on more varied data is likely to be more beneficial when combined with larger models with better generalization capacity.

% ..............................................................................
\subsection{In-distribution} \label{sec:res:base}
We compare our final models to the (self-)supervised \ac{sota} on the two datasets from our training set with available ground-truth, namely Kitti and \acl{mc}.
This represents the evaluation procedure commonly employed by most papers, where the test data is drawn from the same distribution as the training data. 

These results can be found in \tbl{res:all} (\emph{In-Distribution}). 
Both of our models outperform every (self-)supervised baseline on both datasets, excluding NewCRFs~\cite{Weihao2022} on Kitti. 
This shows that \acl{stv} provides complementary driving data that can generalize across datasets. 
Furthermore, the improved \acs{kbr++} increases performance on \acl{mc} due to the additional indoor data provided by \acl{ctv}.

% ..............................................................................
\subsection{Zero-shot Generalization} \label{sec:res:zero}
The core of our evaluation takes place in a \emph{zero-shot} setting, meaning that models are not fine-tuned on the target datasets. 
This tests the capability to adapt to out-of-distribution examples and previously unseen environments. 
Existing \acs{ssmde} publications sometimes include zero-shot evaluations. 
However, this is frequently limited to CityScapes~\cite{Cordts2016} and Make3D~\cite{Saxena2009}, which contain low-quality ground-truth and represent an urban automotive domain similar to the training Kitti. 
We instead opt for a much more challenging collection of datasets, constituting a mixture of urban, natural, synthetic and indoor scenes.
Please refer to \tbl{data:overview} for details regarding the evaluations datasets and their splits.

\heading{Outdoor}
These results can be found in \tbl{res:all} (\emph{Outdoor}).
Both of our models outperform the \ac{ssl} baselines by a large margin. 
This is even the case on DDAD~\cite{Guizilini2020}, which is also an urban automotive dataset.
Meanwhile, our model is capable of generalizing to urban~\cite{Guizilini2020,Vasiljevic2019}, synthetic~\cite{Butler2012} and natural~\cite{Adams2016,Spencer2022} datasets, performing on par with the \ac{sota} supervised baselines. 
It is interesting to note that \acl{crf} generalizes across automotive datasets and provides the best performance on \acs{ddad}.
However, it fails when evaluated in alternative domains and provides only minimal improvements over the \ac{ssl} baselines. 
Finally, even the more efficient \acs{kbr} provides impressive performance that matches more complex transformer-based backbones.

\heading{Indoor}
\tbl{res:all} (\emph{Indoor}) shows performance on all indoor datasets. 
Note that \acl{crf} was trained exclusively on \acl{nyud}, while \acs{dpt-beit} uses it as part of its training collection.
As such, this subset of results is \emph{not} zero-shot.
Our models outperform the \ac{ssl} baselines by an even larger margin, due to the large shift in distribution when moving from outdoor to indoor scenes.
In this case, \acs{kbr++} provides a noticeable improvement over \acs{kbr}, thanks to the additional indoor training data provided by \acl{ctv}.
This helps to further close the gap \wrt supervised approaches. 
Once again, our model is now capable of performing on-par with all supervised models except \acl{dpt-beit}, despite requiring no ground-truth annotations.

\heading{Overall}
To summarize, \tbl{res:all} (\emph{Multi-task}) reports the multi-task metrics across all datasets. 
Our models outperform the updated \ac{ssmde} baselines from~\cite{Spencer2022} by over 35\%.
Meanwhile, the contributions from this paper further improve our original model~\cite{Spencer2023c} by 4.5\%. 
What's more, \acs{kbr++} is the second-best model overall, outperforming supervised baselines such as \acl{crf} and \acl{dpt-vit}.
It is worth emphasizing once again that our model is \emph{entirely self-supervised}, relying exclusively on the photometric consistency across frames.
Thus, we present the first approach demonstrating the true capabilities of \acs{ssmde}, which can leverage much larger and more diverse collections of data freely available on YouTube. 

% =====================================
\begin{landscape}
\thispagestyle{empty}
\begin{table}[htbp]
\footnotesize
\addtolength{\tabcolsep}{-0.2em}
\renewcommand{\arraystretch}{1.2}
\centering
\input{Tables/*}
\label{tbl:[}
\end{table}
t]{res}{res:all}    
\end{landscape}
% =====================================

% =====================================
%
\begin{figure}[htbp]
\centering
\input{Figures/*}
\label{fig:[}
\end{figure}
ht]{viz_zero}{res:viz:zero}
% =====================================

% ..............................................................................
\subsection{Qualitative Results} \label{sec:res:viz}

\heading{Visualizations}
Sample predictions for each model and dataset can be found in \fig{res:viz:zero}.
Our models are significantly more robust than the best \ac{ssl} baseline~\cite{Lyu2021}, which fails on all domains except the automotive. 
This is most noticeable in indoor settings, where it treats human faces as background. 
Meanwhile, our model generalizes across all datasets and environments, providing high-quality predictions. 
It can also be seen how \acs{kbr++} improves over the base \acs{kbr}, especially in thin structures (\acl{dio}) and depth boundaries (\acs{tum}).

\heading{Failure Cases}
Despite being a significant step forward for \acs{ssmde}, our model still has a few limitations and failure cases. 
We show these examples in \fig{res:viz:fail}.
The main one is the lack of explicit modeling for dynamic objects. 
Whilst the minimum reconstruction loss and automasking~\cite{Godard2019} can reduce their impact, there are still cases where vehicles in front of the camera are predicted as holes of infinite depth. 
Another common failure case is texture-copying artifacts, where textures from the original image are incorrectly predicted as changes in depth. 
This can happen on objects such as textured walls or pavements made with bricks or text on shirts and signs. 
Finally, another interesting failure case are reflective or transparent surfaces, as they do not violate the photometric constraints during training. 
However, these are also challenging for supervised methods, as the data cannot be correctly captured using \acs{lidar} either.

% =====================================
%
\begin{figure}[htbp]
\centering
\input{Figures/*}
\label{fig:[}
\end{figure}
t]{viz_fail}{res:viz:fail}
% =====================================

% ..............................................................................
\subsection{Map-Free Relocalization} \label{sec:res:reloc}
Following our preliminary publication~\cite{Spencer2023c}, we report our results on the MapFreeReloc~\cite{Arnold2022} benchmark. 
Map-free relocalization aims to regress the 6DoF pose of a target frame based only on the known pose of a single reference frame. 
This is contrary to traditional localization pipelines, which typically contain a map-building or network-training phase that requires large-scale captures for each specific scene. 
Recent research~\cite{Toft2020,Arnold2022} has shown that the scale ambiguity in map-free relocalization feature-matching pipelines can be resolved by incorporating \ac{sota} \ac{mde} predictions.

The models are evaluated on the validation split of the benchmark, which consists of 37k images from 65 small-scale landmarks. 
The data was crowd-sourced and collected using mobile phones, meaning that it features an uncommon portrait aspect ratio.
This makes the task of zero-shot transfer even more challenging.

The feature-matching baseline~\cite{Arnold2022} uses LoFTR~\cite{Sun2021} correspondences, a PnP solver and DPT~\cite{Ranftl2021} fine-tuned on either Kitti or \acl{nyud}.
Metric depth for all evaluated models is obtained by aligning the predictions to the baseline fine-tuned DPT predictions. 
We report the evaluation metrics provided by the benchmark authors. 
This includes translation (meters), rotation (deg) and reprojection (px) errors.
Pose Precision/AUC were computed with an error threshold of 25 cm \& 5$^\circ$, while Reprojection uses a threshold of 90px.

The results can be found in \tbl{res:mapfree}, along with visualizations in \fig{res:mapfree}.
The updated models from this paper (\acs{kbr++}) outperform every (self-)supervised model, except \acl{dpt-beit}. 
This demonstrates the effectiveness of training with our diverse datasets, as well as the improved robustness to image aspect ratios provided by \ac{araug}.
Furthermore, it showcases the ability to incorporate \acs{ssmde} into real-world problem pipelines. 

We find that the original DPT models perform better than their fine-tuned counterparts, despite using these as the metric scale reference. 
This suggests that the finetuning procedure of~\cite{Arnold2022} may provide metric scale at the cost of generality.  
However, this highlights the need for models that predict accurate metric depth, rather than only relative depth.

% =====================================
%
\begin{table}[htbp]
\footnotesize
\addtolength{\tabcolsep}{-0.2em}
\renewcommand{\arraystretch}{1.2}
\centering
\input{Tables/*}
\label{tbl:[}
\end{table}
t]{mapfree}{res:mapfree}
% =====================================

% =====================================
%
\begin{figure}[htbp]
\centering
\input{Figures/*}
\label{fig:[}
\end{figure}
t]{mapfree}{res:mapfree}
% =====================================

% ------------------------------------------------------------------------------
\section{Conclusion} \label{sec:conc}
This paper has introduced \acs{kbr} and \acs{kbr++}, the first \acs{ssmde} models that match and even outperform \ac{sota} \emph{supervised} algorithms. 
We demonstrated this in our challenging zero-shot experiments, which showcase the robustness and generalization capabilities of our models. 
This was made possible due to our approach to data collection, focusing on the scale of the training set and leveraging the lack of annotations needed for self-supervised learning.
We curated two novel large-scale YouTube datasets, \acl{stv} and \acl{ctv}, with a total of 2M training frames.
These datasets contain an incredibly diverse set of environments, ranging from hikes in snowy forests, to luxurious houses and even underwater caves. 

Performance and generalization were further maximized by introducing stronger augmentation regimes (\ac{araug}, RandAugment and CutOut), simultaneously learning the camera's intrinsics, making the training more flexible and modernizing the network architecture. 
Our extensive ablations demonstrated the benefits of introducing each respective component. 

The main limitation of the current models is their sensitivity to dynamic objects.
Whilst the contributions from Monodepth2~\cite{Godard2019} alleviate some of these artifacts, a more explicit motion model may be required to handle these scenarios. 
Introducing optical flow constraints may be the most feasible way to achieve this in a self-supervised manner.
However, it is worth noting the increased computational requirements resulting from training a new network and computing the required consistency losses.

Finally, estimating metric depth (for generic scenes) without ground-truth annotations is an open research problem that could further increase the applicability of \acs{ssmde} to real-world tasks. 
By making the datasets and code freely available to the public, we hope to further drive the \ac{sota} in \acs{ssmde} and inspire future research that addresses these challenging problems.

% ------------------------------------------------------------------------------
\section*{Acknowledgements}
This work was partially funded by the EPSRC under grant agreements EP/S016317/1 \& EP/S035761/1.

% % ------------------------------------------------------------------------------
% \section*{Declarations}

% Some journals require declarations to be submitted in a standardised format. Please check the Instructions for Authors of the journal to which you are submitting to see if you need to complete this section. If yes, your manuscript must contain the following sections under the heading `Declarations':

% \begin{itemize}
% \item Conflict of interest/Competing interests (check journal-specific guidelines for which heading to use)
% \item Ethics approval 
% \item Consent to participate
% \item Consent for publication
% \item Availability of data and materials
% \item Authors' contributions
% \end{itemize}

% \noindent
% If any of the sections are not relevant to your manuscript, please include the heading and write `Not applicable' for that section. 

% %%===================================================%%
% %% For presentation purpose, we have included        %%
% %% \bigskip command. please ignore this.             %%
% %%===================================================%%
% \bigskip

% Springer journals and proceedings: \url{https://www.springer.com/gp/editorial-policies}

% Nature Portfolio journals: \url{https://www.nature.com/nature-research/editorial-policies}

% \textit{Scientific Reports}: \url{https://www.nature.com/srep/journal-policies/editorial-policies}

% BMC journals: \url{https://www.biomedcentral.com/getpublished/editorial-policies}
% \end{flushleft}

\begin{appendices}

% \section{Section title of first appendix}\label{secA1}

% An appendix contains supplementary information that is not an essential part of the text itself but which may be helpful in providing a more comprehensive understanding of the research problem or it is information that is too cumbersome to be included in the body of the paper.




\end{appendices}


\bibliography{sn-bibliography}% common bib file

\end{document}